\documentclass[letterpaper, 10 pt, conference]{ieeeconf}  

\IEEEoverridecommandlockouts                              

\usepackage{times}
\usepackage{epsfig}
\usepackage{graphicx}
\usepackage{amsmath}
\usepackage{amssymb}
\usepackage{booktabs}
\usepackage{algorithm}
\usepackage{algorithmic}
\usepackage{tablefootnote}
\usepackage{comment}
\usepackage{bbm}  
\usepackage{bm}  
\usepackage{tabu}
\usepackage{xcolor}
\usepackage{caption}
\usepackage{subcaption}
\usepackage{wrapfig}
\usepackage{balance}

\title{\LARGE \bf
PBP: Path-based Trajectory Prediction for Autonomous Driving
}

\author{Sepideh Afshar$^*$, Nachiket Deo$^*$, Akshay Bhagat, Titas Chakraborty, Yunming Shao, \\ Balarama Raju Buddharaju,
Adwait Deshpande, Henggang Cui \\
Motional \\
{\tt\small \{sepideh.afshar, nachiket.deo, henggang.cui\}@motional.com}
  \thanks{
    $^{*}$Authors contributed equally.
  }
}

\begin{document}

\maketitle
\thispagestyle{empty}
\pagestyle{empty}


\begin{abstract}

Trajectory prediction plays a crucial role in the autonomous driving stack by enabling autonomous vehicles to anticipate the motion of surrounding agents. Goal-based prediction models have gained traction in recent years for addressing the multimodal nature of future trajectories. Goal-based prediction models simplify multimodal prediction by first predicting 2D goal locations of agents and then predicting trajectories conditioned on each goal. However, a single 2D goal location serves as a weak inductive bias for predicting the whole trajectory, often leading to poor \textit{map compliance}, i.e., part of the trajectory going off-road or breaking traffic rules. In this paper, we improve upon goal-based prediction by proposing the \emph{Path-based prediction (PBP)} approach. PBP predicts a discrete probability distribution over reference paths in the HD map using the path features and predicts trajectories in the path-relative Frenet frame. We applied the PBP trajectory decoder on top of the HiVT scene encoder and report results on the Argoverse dataset. Our experiments show that PBP achieves competitive performance on the standard trajectory prediction metrics, while significantly outperforming state-of-the-art baselines in terms of map compliance.

\end{abstract}

\section{Introduction}
\label{sec:intro}


To safely navigate through traffic while offering passengers a smooth ride, autonomous vehicles need the ability to predict the trajectories of surrounding agents. 
There is inherent uncertainty in predicting the future, making this a challenging task. Agent trajectories tend to be highly non-linear over long prediction horizons. 
Additionally, the distribution of future trajectories is multimodal; in a given scene an agent could have multiple plausible goals and could take various paths
to each goal.

In spite of these challenges, agent motion is not completely unconstrained. Vehicles tend to follow the direction of motion ascribed to their lanes, make legal turns and lane changes, and stop at stop signs and crosswalks. Bicyclists tend to use the bike lane, and pedestrians tend to walk along sidewalks and crosswalks. High-definition (HD) maps of traffic scenes efficiently represent such constraints on agent motion and have thus been a critical component of autonomous driving datasets \cite{chang2019argoverse, caesar2020nuscenes, ettinger2021large, wilson2021argoverse, caesar2021nuplan}.
In fact, it has been shown in many prior works~\cite{ rhinehart2018r2p2, niedoba2019improving, boulton2020motion, cui2021ellipse, ridel2020scene, greer2021trajectory, zhu2022motion} that a key requirement of the trajectory prediction task for a real-world autonomous driving system is to predict \textit{map-compliant} trajectories -- trajectories that don't go off-road or violate traffic rules over long prediction horizons.
For example, incorrectly predicting a non-map-compliant trajectory that encroaches into the oncoming traffic lane could cause the ego vehicle to brake hard or even make dangerous maneuvers on the road.
As a result, prediction map compliance w.r.t. the provided HD map is central to our proposed approach and experimental evaluation.

Prior works have leveraged HD maps for trajectory prediction in two distinct ways. First, the HD map is often used as an input to the model. Early works \cite{cui2019multimodal, chai2020multipath, phan2020covernet} use rasterized HD maps and CNN encoders. More recent works directly encode vectorized HD maps using PointNet encoders \cite{ye2021tpcn, gao2020vectornet}, graph neural networks \cite{liang2020learning} or transformer layers \cite{ngiam2021scene, liu2021multimodal, nayakanti2022wayformer, zhou2022hivt}. The map encoding is then used by a multimodal prediction header to output $K$ trajectories and their probabilities. A drawback of multimodal prediction headers is that they need to learn a complex one-to-many mapping from the entire scene context to multiple future trajectories, often leading to non-map-compliant predictions. 

\begin{figure*}[t]
    \includegraphics[clip,width=0.9 \linewidth]{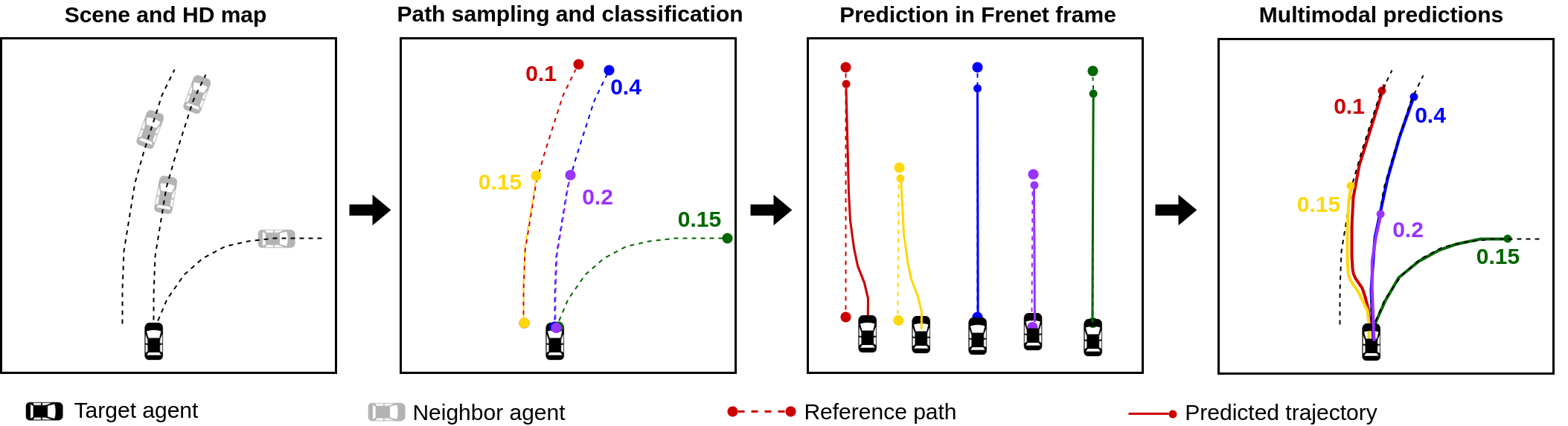}%
    \centering
    \caption{
        \textbf{Overview of path-based prediction.} Path-based prediction predicts trajectories conditioned on \textit{reference paths} rather than 2D goals. We sample reference paths using the lane network from HD maps, predict a discrete distribution over the sampled paths, and predict future trajectories in the Frenet frame relative to the paths. Finally, we transform the trajectories back to the Cartesian frame relative to the target agent to obtain multimodal predictions.
    }
    \label{fig:concept}
\end{figure*}

To address this shortcoming, a few recent works additionally use the HD map for \textit{goal-based prediction} \cite{zhao2021tnt, gu2021densetnt, zeng2021lanercnn, gilles2021home, gilles2022gohome}. Goal-based prediction models associate each mode of the trajectory distribution to a 2D goal location sampled from the HD map. They predict a discrete distribution over the sampled goals, and then predict trajectories conditioned on each goal. This simplifies the mapping learned by the prediction header, and also makes each mode of the trajectory distribution more interpretable. However, 2D goal locations serve as a weak inductive bias to condition predictions, and may lead to imprecise trajectories for each goal.

In this work, we seek to improve upon goal-based trajectory prediction. We argue that \textit{reference paths} rather than 2D goals are the appropriate HD map element to condition predicted trajectories. We define reference paths as segments of lane centerlines close to the agent of interest that the agent may follow over the prediction horizon. We propose a novel path classifier that predicts a discrete probability distribution over the candidate reference paths and a trajectory completion module that predicts trajectories conditioned on each path in the Frenet frame. Figure \ref{fig:concept} shows an overview of our approach. In particular, our approach has two key advantages over goal-based prediction:
\begin{enumerate}
\item \textbf{Path features instead of goal features:} We predict trajectories conditioned on feature descriptors of the entire reference path instead of just 2D goal locations. This is a more informative feature descriptor and leads to more map-compliant trajectories over longer prediction horizons compared to goal-based prediction. 
\item \textbf{Prediction in the Frenet frame:} The reference paths allow us to predict trajectories in the Frenet frame relative to the path. Compared to the Cartesian frame with varying lane locations and curvatures, predictions in the Frenet frame have much lower variance, which leads to more map-compliant trajectories that better generalize to novel scene layouts.    

\end{enumerate}

Our path-based trajectory decoder is modular by design and could be used with any existing scene encoder such as VectorNet \cite{gao2020vectornet}, LaneGCN \cite{liang2020learning}, Scene Transformer \cite{ngiam2021scene}, Wayformer \cite{nayakanti2022wayformer}, etc. Here, we build our decoder on top of the recently proposed HiVT encoder \cite{zhou2022hivt} that achieved competitive results on the Argoverse dataset \cite{chang2019argoverse} and has a publicly available code base. 
Our results on the Argoverse dataset show that our path-based decoder achieves competitive performance in terms of the standard minADE, minFDE, and miss rate metrics, while significantly outperforming the HiVT baseline and goal-based prediction in terms of map compliance metrics.  

Our contributions can be summarized as follows: 
\begin{itemize}
    \item We propose a novel path-based trajectory prediction (PBP) approach that improves upon traditional goal-based prediction.
    \item We applied our PBP trajectory decoder on top of the HiVT~\cite{zhou2022hivt} scene encoder. The resulting model achieves the best map compliance metric on the Argoverse leaderboard while being competitive in terms of prediction error metrics.
    \item We present extensive ablation studies comparing different trajectory decoder approaches on the Argoverse validation set.
\end{itemize}

\begin{figure*}[t]
    \centering
    \includegraphics[clip,width=0.9\linewidth]{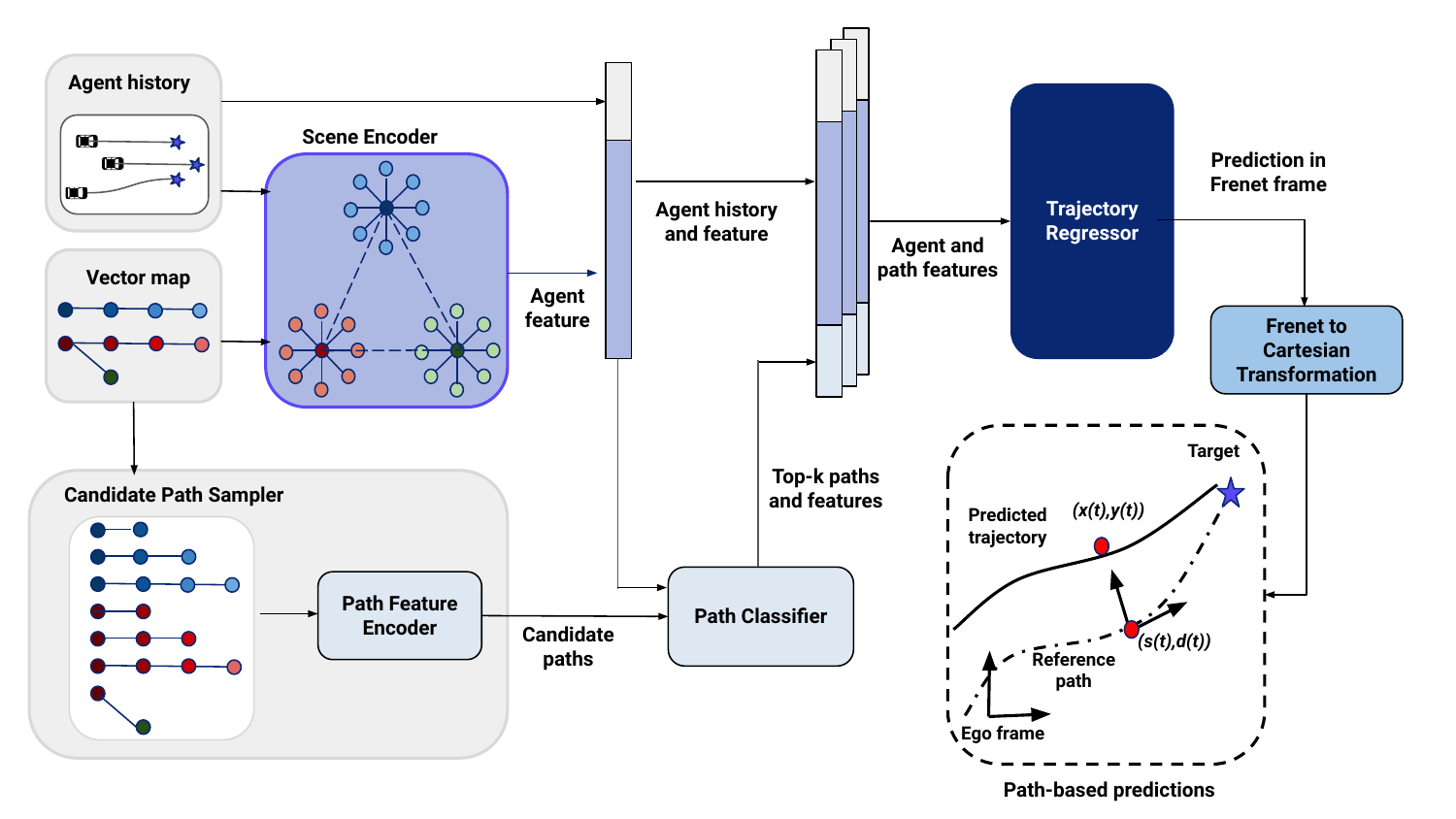}
    \caption{
        \textbf{Model architecture:} Our model consists of four key modules. The \emph{scene encoder} encodes the agent history and HD map information (Section \ref{sec:scene_enc}). The \emph{candidate path sampler} samples candidate paths for each agent from the lane graph (Section \ref{sec:target_sampling}). The \emph{path classifier} predicts a discrete distribution over the reference paths (Section \ref{sec:path_class}). Finally, the \emph{trajectory regressor} decodes trajectory predictions in the path-relative Frenet frame conditioned on the paths (Section \ref{sec:decoder}).
    }
    \label{fig:PathBasedPredModel}
\end{figure*}

\section{Related work}
\label{sec:formatting}


\noindent\textbf{Map-compliant trajectory prediction:} Leveraging the HD-map and predicting map-compliant trajectories has been the focus of a large number of works on trajectory prediction. Several works have proposed novel HD map encoders \cite{djuric2020uncertainty, gao2020vectornet, liang2020learning, ye2021tpcn, ngiam2021scene, zhou2022hivt}, trajectory decoders conditioned on HD maps \cite{zhao2021tnt, gu2021densetnt,gilles2021home, gilles2022gohome, wang2022ltp, deo2022multimodal,zhang2021map}, and even novel metrics and auxiliary loss functions for map-compliance \cite{ rhinehart2018r2p2, niedoba2019improving, boulton2020motion, cui2021ellipse, ridel2020scene, greer2021trajectory, zhu2022motion}.
In this work, we propose a path-based prediction approach that significantly improves prediction map compliance.

\noindent\textbf{Goal-free multimodal prediction:} The distribution of future trajectories is multimodal due to unknown intents of agents. Machine learning models for trajectory prediction thus need to learn a one-to-many mapping from the HD map and past states of agents, to multiple future trajectories. Prior work has addressed this using two approaches. The first approach is to implicitly learn the trajectory distribution using latent variable models such as GANs \cite{gupta2018social, sadeghian2019sophie, zhao2019multi}, CVAEs \cite{lee2017desire, salzmann2020trajectron++}, and normalizing flows \cite{rhinehart2018r2p2, rhinehart2019precog}, where samples from the model represent plausible future trajectories. The other common approach is to use a multimodal regression header that outputs a fixed number of trajectories along with their probabilities \cite{cui2019multimodal, liang2020learning, zhou2022hivt, ngiam2021scene}. Such models are trained using the winner takes all/variety loss \cite{gupta2018social}. Some recent works~\cite{nayakanti2022wayformer, varadarajan2022multipath++, wang2023prophnet}, use DETR-like learned tokens~\cite{carion2020end} to output $K$ distinct trajectories.

\noindent\textbf{Goal-based prediction:} Goal-based prediction models \cite{zhao2021tnt, zeng2021lanercnn, gilles2021home, gilles2022gohome, gu2021densetnt} partly address the above limitations by associating each mode of the trajectory distribution to a 2D goal in the HD map. TNT \cite{zhao2021tnt} samples a sparse set of goals along lane centerlines. LaneRCNN \cite{zeng2021lanercnn} uses nodes in a lane graph to predict goal locations. HOME \cite{gilles2021home} and GoHOME \cite{gilles2022gohome} predict goal heatmaps along a grid and graph representation of the HD map, and sample goal locations to optimize for the minFDE or miss rate metrics. Finally, DenseTNT \cite{gu2021densetnt} first predicts a dense goal heatmap along lanes, before using a second learned model to sample goals from the heatmap. We improve upon goal-based prediction models by conditioning our predictions on reference paths in the HD map rather than goals. Reference paths provide our trajectory decoder with more informative feature descriptors than 2D goal coordinates, and additionally allow us to predict in the path-relative Frenet frame.  

\noindent\textbf{Frenet frame trajectory decoding:} There are some existing models that predict trajectories in path-relative Frenet frame, such as GoalNet~\cite{zhang2021map}, DAC~\cite{narayanan2021divide}, and WIMP~\cite{khandelwal2020if}.
PBP has two key differences from those works.
First, PBP has a different definition of its reference paths from those works.
The reference paths in GoalNet, DAC, and WIMP are fixed-lengthed paths in the lane level.
To generate the reference paths, GoalNet and DAC start from the agent's current position and search along the lane graph for a fixed distance.
Such reference paths only capture the agent's high-level intention (e.g., go straight or turn right) but do not capture other uncertainties such as change of speed profiles.
As a result, GoalNet, DAC, and WIMP all predict $M$ trajectory modes within each reference path to achieve multimodal prediction.
On the other hand, PBP's reference paths are sequences of lane segments with variable lengths, and PBP relies entirely on its path classification to achieve multimodal prediction since a reference path can uniquely define a predictive mode.
To highlight the difference, PBP considers around 600 candidate reference paths per agent, while GoalNet and DAC only consider less than three reference paths per agent.
Second, DAC~\cite{narayanan2021divide} and WIMP~\cite{khandelwal2020if} do not have a learned path classification module to predict path probabilities or a path classification loss as a training objective.
DAC uses a heuristic algorithm to rank paths based on the distance-along-lane score and centerline-yaw score,
and WIMP finds only one single closest reference path for each agent using a heuristic algorithm.
On the other hand, PBP has a path classification module that predicts the probability distribution over all candidate paths.

PRIME~\cite{song2022learning} also predicts trajectories in the Frenet frame, but it uses a model-based trajectory generator (a quartic polynomial) to sample trajectories. In contrast, PBP's trajectory generator is entirely learned, allowing it to generate a variety of motion profiles in the Frenet frame.

\section{PBP: Path-based prediction}

\subsection{Problem statement}
The objective of a trajectory prediction model is to forecast the future trajectories of a set of agents in the scene, given their past history positions and map context. We denote the past history positions of an agent $a$ by $\lbrace\bm{P}^a\rbrace_{Past}=\lbrace \bm{P}^a_{-T'+1}, \bm{P}^a_{-T'+2}, \cdots, \bm{P}^a_{0}\rbrace$ where $\bm{P}^a_t = (x^a_t, y^a_t)$ is a 2-D coordinate position, and $T'>0$ is the past history length.
The map context $\mathcal{M}$ is represented as a set of discretized lane segments $\lbrace l_j \rbrace_{j=1}^L$ and their connections.
The prediction model is required to forecast the future state of each agent $\lbrace\bm{P}^a\rbrace_{Future}=\lbrace \bm{P}^a_{1}, \bm{P}^a_{2}, \cdots, \bm{P}^a_{T}\rbrace$ over the time horizon $T>0$.
In order to capture the uncertainties of the agents' future behaviors, the model will output $K$ trajectory predictions and their probabilities $\lbrace p_k \rbrace_{k=1}^K$ for each agent.

\subsection{Overall architecture}

The overall architecture of our PBP model is illustrated in Figure \ref{fig:PathBasedPredModel},
which consists of four main components.
The scene encoder generates agent and map embeddings from agent-map and agent-agent interactions (Section \ref{sec:scene_enc}).
The candidate path sampler samples the candidate paths from the map for each agent (Section \ref{sec:target_sampling}).
The path classifier predicts the probability of each sampled path (Section \ref{sec:path_class}).
Finally, the trajectory regressor decodes trajectories conditioned on the selected paths (Section \ref{sec:decoder}).

\subsection{Scene encoding}
\label{sec:scene_enc}
The scene encoder module creates agent feature vectors from the scene for each agent.
In this work, we borrowed the scene encoder module from the HiVT model \cite{zhou2022hivt}, a recently proposed trajectory prediction model that achieves state-of-the-art performance on Argoverse.
The HiVT scene encoder represents each scene as a set of vectorized entities. It uses this representation to encode the scene by hierarchical aggregation of the spatial-temporal information.
First, rotational invariant local feature vectors are encoded for each agent with a transformer module to aggregate neighboring agents’ information as well as local map structure. Next, global interactions between agents are aggregated into each agent's feature vector to capture the scene-level context.
The outputs of the encoder are the feature vectors for each agent denoted by $\bf{F}_a$.

\subsection{Candidate sampling}
\label{sec:target_sampling}
The objective of the candidate sampling module is to create a set of candidate reference paths for each agent by traversing the lane graph.
A reference path is defined as a sequence of connected lane segments $r_i=\{l_{i,1}, l_{i,2}, \cdots, l_{i,R_i}\}$.
The starting point of the reference path for an agent $a$ is supposed to be in the vicinity of the agent's current location $\bm{P}^a_0$, and the endpoint is supposed to be in the vicinity of the agent's future trajectory endpoint $\bm{P}^a_T$, as is illustrated in Figure~\ref{fig:concept}.

To select the candidate reference path for an agent $a$, we first select a set of \emph{seed lane segments} that will be considered as the path starting points.
We used a simple heuristic to select the seed lane segments by picking the lane segments that are within a distance range of the agent's current location and have their lane directions within a range of the agent's current heading.
By picking the seed lanes this way, we will have candidate paths starting from not only the agent's current lane but also the neighbor lanes, which allows the model to predict lane-changing trajectories.


From the seed lane segments, we run a breadth-first search to find the candidate paths. The output of the candidate sampling module is a set of candidate reference paths for each agent, denoted as $\mathcal{R}^a=\lbrace r_i^a \rbrace$.


\subsection{Path classification}
\label{sec:path_class}
Given the set of candidate reference paths, the path classification module predicts the probability distribution over them using the agent and path features.

To encode the features $\bm{F}_{p,i}$ of a path $r_i=\{l_{i,1}, l_{i,2}, \cdots, l_{i,R_i}\}$, we pick the the start segment $l_{i,1}$, the middle segment $l_{i,R_i // 2}$, and the end segment $l_{i,R_i}$ of the path,
and use their coordinates and direction vectors as the raw feature.
We encode those raw features with an MLP to a feature vector $\bm{F}_p$.

In addition to the agent and path features, we also create an agent-path pair feature that captures the interactions between the agent and the path.
We use the distance vectors and angle deltas from the agent's current location to the start, middle, and end segments of the path as the raw features. We then use another MLP network to encode them to an agent-path pair feature vector $\bm{F}_{a, (p,i)}$


We concatenate the agent feature $\bm{F}_a$, path feature $\bm{F}_p$, and agent-path pair feature $\bm{F}_{a, (p,i)}$ together and run them through another MLP network to predict the probability distribution over all candidate paths of the agent, trained with the cross-entropy loss as $\mathcal{L}_{cls}$.
We decide the ground-truth reference path $r^a_{GT}$ of the agent $a$ based on its ground-truth future trajectory $\lbrace\bm{P}^a\rbrace_{Future}$, similar to the ground-truth goal selection in goal-based prediction.
At inference time, we use non-maximum suppression (NMS) to sample a set of $K$ diverse paths to decode the trajectory predictions.


\begin{table*}[]
\centering
\caption{Decoder ablations on Argoverse validation set.
}
\begin{tabular}{@{}lcccccc@{}}
\toprule
Decoder                           & minFDE$_1$ & MR$_1$  & minFDE$_6$ & MR$_6$ & \multicolumn{1}{c}{\begin{tabular}[c]{@{}c@{}}Offroad \\ rate\end{tabular}} & \multicolumn{1}{c}{\begin{tabular}[c]{@{}c@{}}Lane \\ dev.\end{tabular}} \\ \midrule
Multimodal regression             & 2.93        & 0.481    & \textbf{0.996}        & 0.101    &  0.069   & 0.510                                                                   \\
Anchor-based           &  2.93       & 0.491    &  1.019       & 0.096    & 0.068    &  0.503                                                                  \\
Goal-based              & \textbf{2.82}        & 0.488    & 1.095        &  0.107   &  0.008   & \textbf{0.386}                                                                  \\
\midrule
PBP in Cartesian frame      & 2.84        & 0.479    & 1.048        & 0.099    & 0.005    &  0.389                                                                 \\ 
PBP (Ours)                        & \textbf{2.82}        & \textbf{0.473}     & 1.008        & \textbf{0.095}    & \textbf{0.004}    & \textbf{0.386}                                                                  \\ \bottomrule
\end{tabular}
\label{tab:ablation}
\end{table*}

\subsection{Frenet frame trajectory decoding}
\label{sec:decoder}
The trajectory regressor module decodes trajectories conditioned on the reference paths.
One key difference between our trajectory regressor and the one used in traditional goal-based prediction~\cite{zhao2021tnt, zeng2021lanercnn, gilles2021home, gilles2022gohome, gu2021densetnt} is that it has the information of the whole reference path instead of just the final goal endpoint.
To leverage this path information, we designed our trajectory regressor to decode trajectories in the path-relative Frenet frame.


For each selected reference path $r^a_i$, the trajectory regressor predicts a trajectory in path-relative Frenet frame, with longitudinal component $\lbrace\hat{s}^a_t\rbrace_{t=1 \cdots T}$ and lateral component $\lbrace\hat{d}^a_t\rbrace_{t=1 \cdots T}$, whose inputs include agent features $\bm{F}_a$, path features $\bm{F}_{p, i}$, and agent history in Frenet frame $\bm{P}^a_{Past, r^a_i}$.

During training, we use a teacher-forcing technique and train the trajectory regressor using the ground-truth reference path $r^a_{GT}$.
We transform the ground-truth trajectory $\bm{P}^a_{Future}$ to the Frenet frame w.r.t. $r^a_{GT}$, with longitudinal component $\lbrace s^a_t\rbrace_{t=1 \cdots T}$ and lateral component $\lbrace d^a_t\rbrace_{t=1 \cdots T}$.

The loss function is defined as smooth $L1$ losses of the longitudinal and lateral components in the Frenet frame:
\begin{equation}
    \label{loss_reg}
    \mathcal{L}^a_{reg}=\sum_{t=1}^T \mathcal{L}_{L1}(s^a_t, \hat{s}^a_t) + \lambda_{lateral} \mathcal{L}_{L1}(d^a_t, \hat{d}^a_t)
\end{equation}

The total loss is a weighted sum of the path classification loss and the trajectory regression loss over all agents:
\begin{equation}
\mathcal{L}_{pbp}=\sum_{a\in \text{Agents}} \lambda_{cls}\mathcal{L}^a_{cls} + \mathcal{L}^a_{reg}
\end{equation}

After predicting the trajectories in the Frenet frame, we transform them back to the Cartesian frame using the corresponding reference path, using the formulas in~\cite{werling2010optimal}.

\subsection{Path-free prediction for non-map-compliant agents}

In order to robustly handle non-map-compliant agents (i.e., agents whose behaviors are not compliant with the annotated map), we additionally train a path-free trajectory decoder with the same architecture as the original HiVT decoder~\cite{zhou2022hivt}.
We also train a binary classifier to select the predictions between the two decoders for each agent.
The path-free decoder and its classifier share the same scene encoder as the PBP decoder and use the agent feature vector $\bm{F}_a$ as the input.
During training, we label an agent as a path-free agent if its ground-truth trajectory is more than 5 meters away from any candidate reference path.

\begin{table*}[]
\centering
\caption{Comparison to the state-of-the-art models on the Argoverse leaderboard}
\begin{tabular}{@{}lccccccc@{}}
\toprule
Model             & minADE$_1$ & minFDE$_1$ & MR$_1$   & minADE$_6$ & minFDE$_6$ & MR$_6$   & DAC  \\ \midrule
TNT \cite{zhao2021tnt}      & 2.174   & 4.959   & 0.710 & 0.910  & 1.446   & 0.166   &   0.9889     \\
DenseTNT \cite{gu2021densetnt}      & 1.679   & 3.632   & 0.584 & 0.882  & 1.282   & 0.126   &   0.9875     \\
GoHOME \cite{gilles2022gohome}      & 1.689   & 3.647   & 0.572 & 0.943  & 1.450   & 0.105   &   0.9811     \\
PRIME \cite{song2022learning} & 1.911   & 3.822   & 0.587 & 1.219  & 1.558   & 0.115   &   0.9898     \\
HiVT-128 \cite{zhou2022hivt}         & 1.598   & 3.532   & 0.547 & 0.773   & 1.169   & 0.127   &  0.9888     \\
MultiPath++ \cite{varadarajan2022multipath++}      & 1.623   & 3.614   & 0.564 & 0.790   & 1.214   & 0.132   &   0.9876    \\
DCMS \cite{ye2022dcms}             & \textbf{1.477}   & \textbf{3.251}   & \textbf{0.532} & 0.766   & 1.135   & 0.109      &   0.9902  \\
Wayformer \cite{nayakanti2022wayformer}        & 1.636   & 3.656   & 0.572 & 0.767   & 1.162   & 0.119      &  0.9893  \\
QCNet \cite{zhou2023query}        & 1.523   & 3.342   & 0.526 & \textbf{0.734}   & \textbf{1.067}   & \textbf{0.106}      &  0.9887  \\ \midrule
PBP (Ours) &  1.626       &   3.562      &  0.535     &   0.855      &  1.325       &  0.145     & \textbf{0.9930}               \\ \bottomrule
\end{tabular}
\label{tab:leaderboard}
\end{table*}

\section{Experiments}
\label{sec:exp}

\subsection{Dataset}
We evaluate our model using the public Argoverse dataset~\cite{chang2019argoverse}. Argoverse includes track histories of agents published at 10 Hz and vectorized HD maps.
The task involves predicting the future trajectory of a focal agent in each scenario over a prediction horizon of 3 seconds, conditioned on 2 seconds of track histories and the HD map of the scene. 

\subsection{Implementation details}
We implemented our path-based prediction decoder on top of the open-source HiVT-64 scene encoder~\cite{zhou2022hivt}. We followed a similar training scheme as the original HiVT model for PBP and its variants.
We used 8 AWS T4 GPUs for model training and evaluation. We trained each model for 64 epochs with a batch size of 4 and the Adam optimizer with a learning rate of 0.0005 and a decay weight of 0.0001.


\subsection{Metrics} 

\noindent\textbf{Best-of-K metrics:} We report results using the standard metrics used for multimodal trajectory prediction: minADE$_K$, minFDE$_K$ and miss rate (MR$_K$). The standard metrics compute prediction errors using the best of $K$ predicted trajectories, in order to not penalize diverse but plausible modes predicted by the model. The minADE$_K$ metric averages the L2 norms of displacement errors between the ground truth and the best mode over the prediction horizon. The minFDE$_K$ metric computes the L2 norm of the displacement error between the final predicted waypoint of the best mode and the final waypoint in the ground truth. Finally, miss rate computes the fraction of all predictions where none of the $K$ predicted trajectories are within 2 meters of the ground truth. We report results for $K$=1 and $K$=6, following the convention used in Argoverse.


\noindent \textbf{Map compliance metrics:}
A key limitation of the standard best-of-k metrics is that they fail to penalize implausible predictions, even if they veer off-road or violate lane directions. Ideally, we want all $K$ predictions to be plausible and map-compliant. Thus, we additionally report two map-compliance metrics.
\textit{Offroad rate} measures the fraction of the predicted waypoints at a given horizon falling outside the drivable area. This is closely related to Argoverse's drivable area compliance (DAC) metric, but our offroad rate metric measures each individual waypoint and can report map compliance as a function of the prediction horizon as in Figure~\ref{fig:offroad_rate}.
\textit{Lane deviation} measures the L2 distance between a predicted waypoint and the nearest lane centerline. It captures map compliance signals even when the waypoint is inside the drivable area.
We report the two map-compliance metrics averaged over all waypoints along the whole prediction horizon and all $K=6$ trajectories.

\subsection{Decoder ablation study}
\label{sec:ablation}

We first perform a set of controlled experiments comparing our PBP model with path classification and Frenet frame trajectory decoder against the following alternative prediction decoders.  

\begin{itemize}
    \item \textit{Multimodal regression:} This is the original HiVT-64 model \cite{zhou2022hivt}. It directly regresses multimodal predictions with the winner-takes-all loss. 

    \item \textit{Anchor-based:} This decoder is used in MultiPath~\cite{chai2020multipath}. It predicts offsets with respect to fixed anchor trajectories. We obtain the anchors using K-means clustering on the train set. 
    
    \item \textit{Goal-based:} The goal-based prediction decoder~\cite{zhao2021tnt, zeng2021lanercnn, gu2021densetnt} uses only the goal endpoint features (no path features) in its goal classification module and decodes trajectories conditioned on goal endpoints (no Frenet frame).
    
    \item \textit{PBP in Cartesian frame:} This decoder performs path classification as in PBP but decodes trajectories in the Cartesian frame instead of the Frenet frame.
    
\end{itemize}

\begin{figure}
    \centering
    \includegraphics[width=0.8\columnwidth]{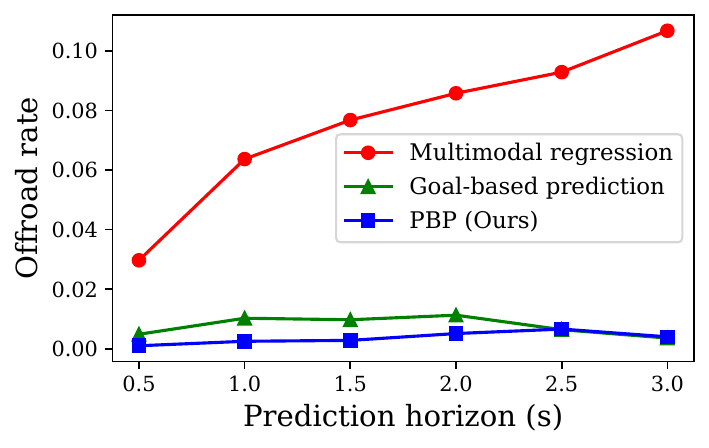}
    \caption{Offroad rate.}
    \label{fig:offroad_rate}
\end{figure}

For fair comparisons, we implemented all decoders using the same HiVT-64 encoder as PBP. The results are shown in Table~\ref{tab:ablation}, and we observe the following.

\vspace{0.1in}
\noindent\textbf{Significantly better map compliance.}
PBP and goal-based prediction achieve significantly lower offroad rates and lane deviation errors than multimodal regression and anchor-based decoders.
This effect is even more pronounced over longer prediction horizons, as shown in Figure~\ref{fig:offroad_rate}.

\noindent \textbf{Advantage over goal-based prediction.} Compared to goal-based prediction, PBP achieves overall lower prediction errors in terms of minFDE and MR and better map compliance metrics, because of the usage of richer path features.
From Figure~\ref{fig:offroad_rate}, goal-based prediction has strong map compliance at the final waypoint (i.e., goal endpoint), but it has higher offroad rates at the intermediate waypoints than PBP because of the missing path information.

\begin{figure}
     \centering
     \begin{subfigure}[b]{0.23\textwidth}
         \centering
         \includegraphics[width=\textwidth,trim={2in 1.5in 1in 1.5in},clip]{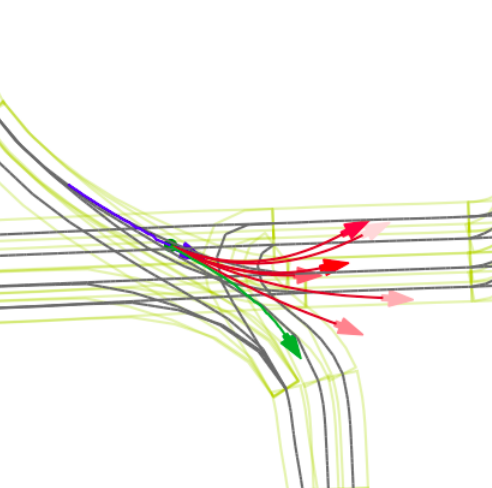}
     \end{subfigure}
     \begin{subfigure}[b]{0.23\textwidth}
         \centering
         \includegraphics[width=\textwidth,trim={2in 1.5in 1in 1.5in},clip]{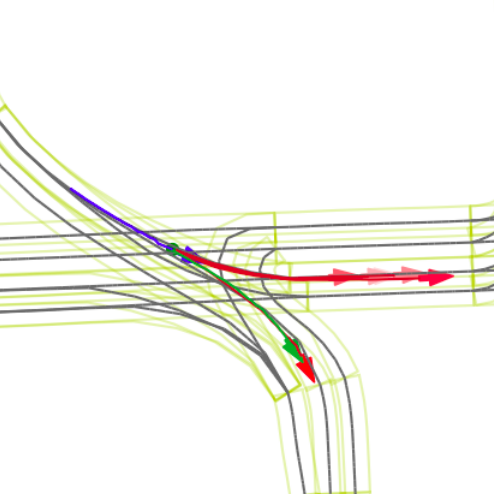}
     \end{subfigure}
     \begin{subfigure}[b]{0.23\textwidth}
         \centering
         \includegraphics[width=\textwidth,trim={1in 1in 2in 2.5in},clip]{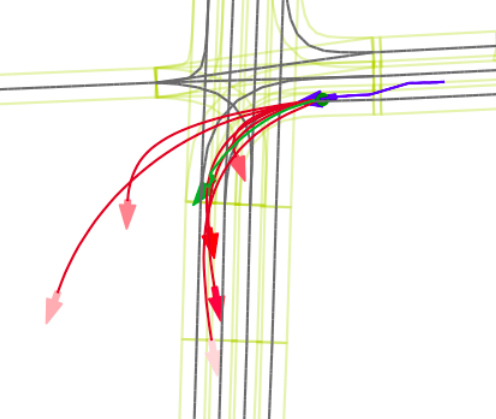}
     \end{subfigure}
     \begin{subfigure}[b]{0.23\textwidth}
         \centering
         \includegraphics[width=\textwidth,trim={1in 1in 2in 2.5in},clip]{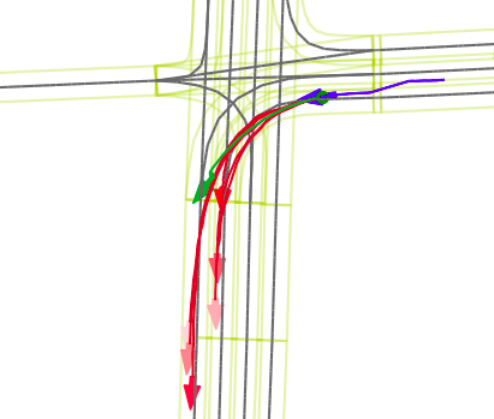}
     \end{subfigure}
     \begin{subfigure}[b]{0.20\textwidth}
         \centering
         \includegraphics[width=\textwidth,trim={2in 1in 0.5in 1in},clip]{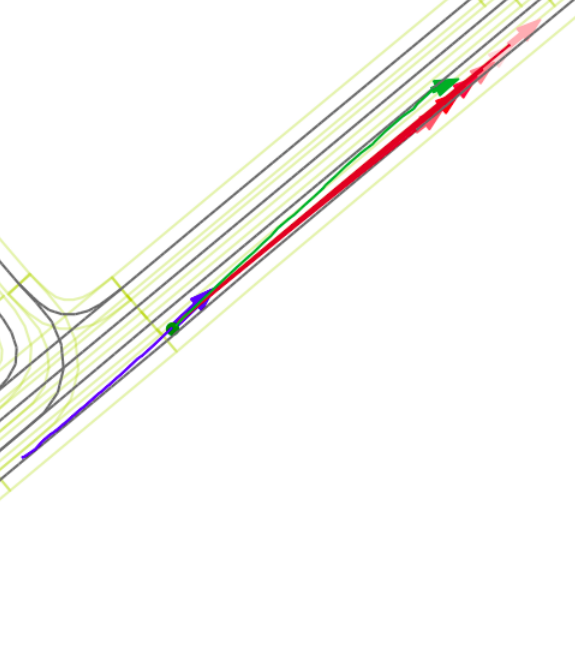}
         \vspace{-0.5in}
         \caption{HiVT-64}
     \end{subfigure}
     \begin{subfigure}[b]{0.20\textwidth}
         \centering
         \includegraphics[width=\textwidth,trim={2in 1in 0.5in 1in},clip]{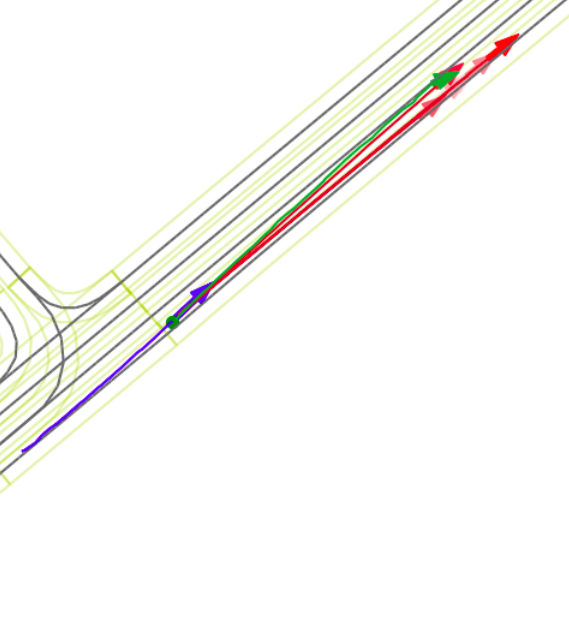}
         \vspace{-0.5in}
         \caption{PBP}
     \end{subfigure}
        \caption{\textbf{Qualitative comparison between original \mbox{HiVT-64} and PBP.} The first column shows the predictions from HiVT-64, and the second column shows the predictions from PBP. The blue, green, and red lines represent past history, ground-truth, and \mbox{top-6} prediction trajectories, respectively.}
        \label{fig:qual}
\end{figure}

\noindent\textbf{Slightly worse mode diversity than goal-free decoders.} PBP's minFDE$_6$ metric is slightly worse than the multimodal regression baseline by 1\%.
This lower diversity is because PBP's predictions are constrained to lanes (as is shown in Figure~\ref{fig:qual}).
We argue that it is a fair trade-off to have more map-compliant predictions for real-world autonomous driving applications.

\subsection{Comparison against the state-of-the-art}

We submitted our PBP model to the Argoverse leaderboard. Table \ref{tab:leaderboard} reports our results along with the top entries on the leaderboard. Our model achieves the highest drivable area compliance (DAC) on the leaderboard, outperforming state-of-the-art in terms of map compliance, while being competitive in terms of minADE$_1$, minFDE$_1$, and MR$_1$.
Those results are consistent with our ablation study results on the validation set.
PBP's top-$6$ metrics are slightly worse than the top leaderboard submissions, but note that most of them used extensive model ensembling (e.g.,~\cite{nayakanti2022wayformer, varadarajan2022multipath++, zhou2023query, wang2022tenet, ye2022dcms}), while our submission used only one single pair of encoder and decoder.
Our inference latency is 72.7 $ms$ on an AWS T4 GPU, with 43.0 $ms$ on the scene encoder and 29.7 $ms$ on the trajectory decoder.



\subsection{Qualitative examples}


Figure~\ref{fig:qual} shows a few qualitative comparisons between the HiVT-64 baseline (using multimodal regression) and PBP.
The results show PBP predicts map-compliant trajectories from all modes, while HiVT-64 has many offroad predictions.
The example on the last row shows that PBP is able to correctly predict lane-changing trajectories because the path candidates also contain paths on the neighbor lanes.

\section{Conclusion}

In this paper, we propose PBP, a novel path-based prediction approach.
In contrast to the traditional goal-based prediction approaches, PBP performs classification on the whole reference path instead of just the goal endpoint.
The additional reference path information improves the path classification accuracy and allows PBP to decode trajectories in the path-relative Frenet frame.
Evaluation results show that the path-based prediction approach makes the trajectory predictions significantly more map-compliant compared to the traditional multimodal regression and goal-based prediction approaches, while maintaining competitive or better prediction accuracy.





\newpage

\bibliographystyle{IEEEtran}
\bibliography{root}

\balance

\end{document}